\documentclass{article}

\usepackage[preprint]{neurips_2025}

\usepackage[utf8]{inputenc} 
\usepackage[T1]{fontenc}    
\usepackage{hyperref}       
\usepackage{url}            
\usepackage{booktabs}       
\usepackage{amsfonts}       
\usepackage{nicefrac}       
\usepackage{microtype}      
\usepackage{xcolor}         
\usepackage{amsmath}
\usepackage{enumitem}
\usepackage{graphicx}
\usepackage{colortbl}  
\usepackage{subcaption}
\usepackage{amsthm}

\theoremstyle{plain}

\theoremstyle{definition}

\theoremstyle{remark}

\title{Self-Regulating Cars: Automating Traffic Control in Free Flow Road Networks}

\author{%
  Ankit Bhardwaj \\
  Department of Computer Science\\
  New York University\\
  New York, USA, 10011 \\
  \texttt{bhardwaj.ankit@nyu.edu} \\
  \And
  Rohail Asim \\
  Department of Computer Science\\
  New York University\\
  Abu Dhabi, UAE, 129188 \\
  \texttt{rohail.asim@nyu.edu} \\
  \AND
  Sachin Chauhan \\
  Department of Computer Science\\
  Indian Institute of Technology Delhi\\
  New Delhi, India 110016 \\
  \texttt{sachin@cse.iitd.ac.in} \\
  \And
  Yasir Zaki \\
  Department of Computer Science\\
  New York University\\
  Abu Dhabi, UAE, 129188 \\
  \texttt{yasir.zaki@nyu.edu} \\
  \And
  Lakshminarayanan Subramanian \\
  Department of Computer Science\\
  New York University\\
  New York, USA, 10011 \\
  \texttt{lakshmi@nyu.edu} \\
}

\begin{document}

\maketitle

\begin{abstract}

Free-flow road networks, such as suburban highways, are increasingly experiencing traffic congestion due to growing commuter inflow and limited infrastructure. Traditional control mechanisms—traffic signals or local heuristics—are ineffective or infeasible in these high-speed, signal-free environments. We introduce self-regulating cars, a reinforcement learning-based traffic control protocol that dynamically modulates vehicle speeds to optimize throughput and prevent congestion, without requiring new physical infrastructure. Our approach integrates classical traffic flow theory, gap acceptance models, and microscopic simulation into a physics-informed RL framework. By abstracting roads into super-segments, the agent captures emergent flow dynamics and learns robust speed modulation policies from instantaneous traffic observations. Evaluated in the high-fidelity PTV Vissim simulator on a real-world highway network, our method improves total throughput by 5\%, reduces average delay by 13\%, and decreases total stops by 3\% compared to the no-control setting. It also achieves smoother, congestion-resistant flow while generalizing across varied traffic patterns—demonstrating its potential for scalable, ML-driven traffic management.

\end{abstract}

\section{Introduction}
\label{sec:intro}

Traffic congestion leads to significant economic and productivity losses globally \citep{badger2013traffic,time_jam}, and now increasingly affects high-speed freeways due to suburban sprawl and commuter traffic flows \citep{gordon1996traffic}. While infrastructure expansion is prohibitively expensive, improved vehicle-level control offers a scalable alternative.
Leveraging the rise of AI-driven vehicles, we introduce self-regulating cars—an automated traffic management system where smart vehicles adjust their speeds using a centralized, reinforcement learning-based protocol. This system enables dynamic, infrastructure-free congestion control by coordinating individual vehicle speeds to maintain high throughput and prevent traffic jams.

We frame freeway congestion control as a centralized reinforcement learning (RL) problem, where an agent dynamically sets segment-level speed limits to optimize network throughput. Unlike traditional traffic signaling protocols, which rely on distributed, reactive heuristics at intersections, our method learns a global policy that coordinates across the network while accounting for flow-density dynamics and stochastic inflow. We incorporate domain knowledge from traffic flow theory, gap acceptance models, and car-following dynamics into the RL design: key traffic metrics (e.g., density, speed, inflow) form the state space, and the reward penalizes congestion while promoting high-speed flow. The agent operates in a discretized action space to support human-in-the-loop deployment and actuation robustness. We train and evaluate our system using the PTV Vissim microscopic simulator on a real-world free-flow road network in Mainz, Germany. Compared to baseline signaling protocols, our approach consistently achieves higher throughput, smoother flow, and greater resilience under varied traffic conditions.

A central challenge in our setting is maintaining smooth, uninterrupted flow in the presence of stochastic traffic inflows and dynamic merging behavior. Unlike signalized traffic control, where stop-and-go behavior is imposed by design, free-flow networks are vulnerable to spontaneous congestion waves that emerge from local interactions, even without explicit bottlenecks. Our RL agent must therefore learn a delicate balance: modulating speeds enough to prevent congestion, but not so aggressively as to induce braking cascades or shockwaves. To enable this, we aggregate vehicle dynamics over larger “super-segments” to capture emergent flow behaviors while smoothing over short-term fluctuations. The discrete action space reflects real-world deployability constraints, enabling implementation via in-vehicle apps or interfaces, while the use of instantaneous state representations—rather than temporal features—supports generalization across varied and stochastic traffic patterns. This design ensures the agent learns policies that are physically grounded and robust to input shifts, rather than overfitting to specific temporal trajectories.

In this paper, we address these challenges through the following key contributions:

\begin{itemize}[noitemsep,leftmargin=*,topsep=0pt]
\item We formulate freeway traffic flow regulation as a centralized reinforcement learning problem aimed at maximizing network throughput while maintaining smooth, congestion-free operation in stochastic, free-flow settings.

\item We design a physics-informed RL framework that incorporates traffic-theoretic insights into both state and reward structures, enabling robust and interpretable policy learning without temporal input dependencies.

\item We introduce a super-segment abstraction to capture emergent flow dynamics and mitigate local noise, and leverage a discretized action space to support human-in-the-loop deployment via in-vehicle interfaces.

\item We evaluate our protocol in the high-fidelity PTV Vissim simulator on a real-world road network under diverse traffic conditions. Compared to baseline distributed control strategies, our approach improves network throughput by 5\%, reduces average vehicle delay by 13\%, and lowers total stops by up to 5\%, while maintaining smooth and robust traffic flow.
\end{itemize}

Taken together, our results highlight a novel and practical traffic management paradigm: learning global speed modulation policies that achieve system-level coordination without infrastructure changes. By aligning with the capabilities of autonomous vehicles and real-time routing platforms, our method offers a feasible path to deployment through lightweight, app-level integration. Large-scale Mobile telematics platforms like CMT\citep{cmt2025} and Samsara\citep{samsara2025} are already adopted by tens of millions of drivers receiving instantaneous coordination and feedback from cloud-AI powered telematics systems, thereby  underscoring the deployability of our approach and bridging ML innovation with scalable impact in modern traffic systems.

\section{Related Work}
\label{sec:related}

The problem of road traffic management, especially freeway traffic management has been approached from many different angles across several research communities.

Traditional freeway management techniques aim to optimize flow, reduce congestion, and enhance safety through methods such as ramp metering \citep{arnold1998ramp,papageorgiou2002freeway,shaaban2016literature}, lane metering \citep{karimi2019optimal,hussain2016freeway}, and dynamic tolling \citep{mattsson2008road,croci2016urban,yang2020marginal,may2000effects}. These infrastructure-heavy methods coordinate traffic through signaling, lane controls, and pricing strategies, but have proven inadequate in handling growing suburban congestion \citep{gordon1996traffic}. In contrast, more recent traffic management proposals fall into two broad categories: network-based approaches \citep{path1993,pems,bhp,tti,mctrans}, which modify physical infrastructure, and vehicle-based approaches like Uber and Google Maps \citep{uber_routing,google_routing}, which provide real-time routing feedback. However, existing vehicle-based systems focus on individual optimization rather than system-level coordination. Our work addresses this gap by introducing a vehicle-based traffic control framework that learns global policies for network-wide regulation, without requiring new infrastructure.

Speed modulation as a method of road traffic management has been previously studied in \citep{oh2005dynamic,soriguera2013assessment}. \citet{oh2005dynamic} use traditional theoretical models of free-flow traffic and derive the optimal speed control mechanism, but do not test the performance of their model with either empirical or simulation experiments. \citet{soriguera2013assessment}, on the other hand, assess the impact of a particular case study in Barcelona, and do not propose any optimal strategy. 

Prior works have also tackled the problem of road traffic management using reinforcement learning \citep{prabuchandran2014multi,wiering2000multi,chu2019multi,li2021network,chauhan2020ecolight,chauhan2023reallight,chauhan2024frugallight}, but most of the works in this domain apply reinforcement learning for learning optimal traffic signal controls, and do not apply to freeways. In the context of free-flow road networks, the applications of reinforcement learning have been studied mostly with cooperative vehicular networks \citep{pan2021integrated,peng2021connected,shi2021connected,vrbanic2021reinforcement}, which employ vehicle-level speed control and lane management strategies for traffic flow management. However, such approaches run into scalability issues since the multi-agent reinforcement learning model is dependent on peer-to-peer communication that scales super-linearly with the number of vehicles. Our system addresses this very important limitation, bringing our evaluation scale close to real-world deployment.

Finally, we also refer to smart city-scale traffic signaling systems like SCATS and SCOOT \citep{stevanovic2009,scats,scoot,Lowrie1982TheSC,Hunt1981SCOOTaTR}, which we initially planned to use as baselines for comparison, but are proprietary protocols that require external licenses that we were unable to procure.

\section{Preliminaries}
\label{sec:preliminaries}
In this section, we introduce concepts that play a key role in the design of our reinforcement learning system.

\paragraph{Traffic Flow Theory}
\label{sec:traffic_flow_theory}
We focus on free-flow road networks, which lack traffic signals and operate under a maximum speed limit. For a single free-flow segment, traffic dynamics follow the volume-density relationship \( C = b \cdot v \), where \( C \) is flow volume, \( b \) is density, and \( v \) is velocity. Unlike traditional fluid models, traffic flow accounts for compressibility (variable density) and collision avoidance. The speed-density relationship is modeled via formulations by \citet{greenshields1935study}, \citet{underwood1961speed}, and \citet{drake1967statistical}, all of which yield a unimodal volume-density curve with a maximum at a critical density—validated empirically in \citet{hall1986empirical}. In multi-segment networks with merges and crossings, conflicting streams complicate the dynamics. These are typically resolved either via signals or through gap acceptance models \citep{traffic_curve}, such as those by \citet{troutbeck1996unsignalized}, \citet{mcdonald1978capacity}, \citet{tanner1962theoretical}, and \citet{harders1968capacity} for unsignalized intersections. These models assume that drivers assess the safety of proceeding by identifying a time-based "gap" in traffic—measured as the interval between vehicles. While effective conceptually, such models rely on simplified geometric assumptions and become analytically intractable under complex real-world intersection conditions.

\paragraph{Traffic Simulation Framework}
Theoretical models of traffic flow are encoded into simulators for studying complex real-world traffic behavior. Macroscopic simulators \citep{papageorgiou2010traffic, ptvvisum} are generally used for studying aggregated variables such as density, flow, and speed at scale, using continuum traffic flow mechanics \citep{lighthill1955kinematic,richards1956shock,payne1973freeway,whitham2011linear}. Microscopic simulators, in contrast, simulate individual vehicles governed by rule-based car-following models \citep{hunter2021vissim,weidmann1993transporttechnology,newell1961nonlinear}. This approach suits our RL-based setup, where traffic flow is modulated at the vehicle level. For our simulations, we use PTV Vissim~\citep{ptvvissim}, a state-of-the-art microscopic traffic simulator employing the Weidmann 99 driving model~\citep{hunter2021vissim}, a psycho-physical behavior model where drivers adjust actions based on discrete thresholds of gap, speed, and relative speed. The key Weidmann parameters were tuned to reproduce realistic speed-density-flow relationships as showin in Figure \ref{fig:poc} (top-left), explained in detail in Appendix \ref{sec:tuning}.

\begin{figure*}[t]
  \centering
  \begin{subfigure}[b]{0.4\linewidth}
    \centering
    \includegraphics[width=\linewidth]{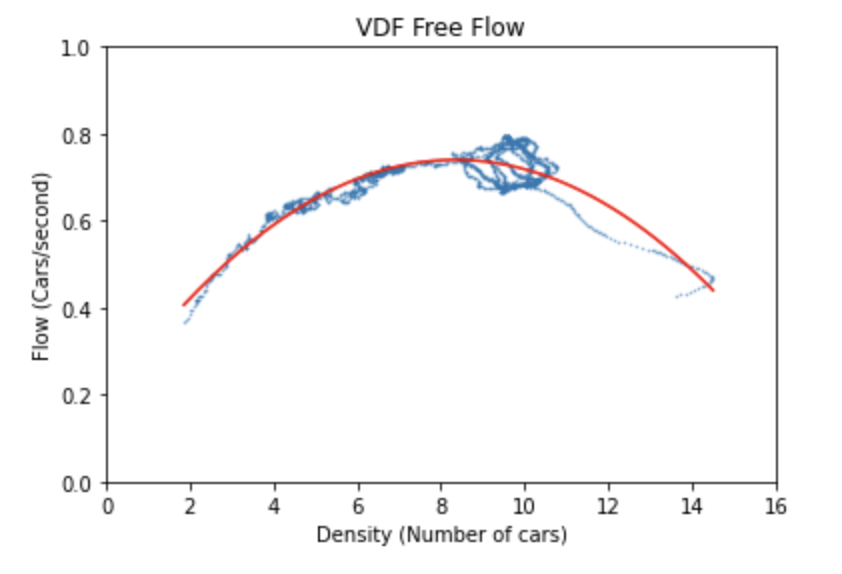}
  \end{subfigure}
  \begin{subfigure}[b]{0.4\linewidth}
    \centering
    \includegraphics[width=\linewidth]{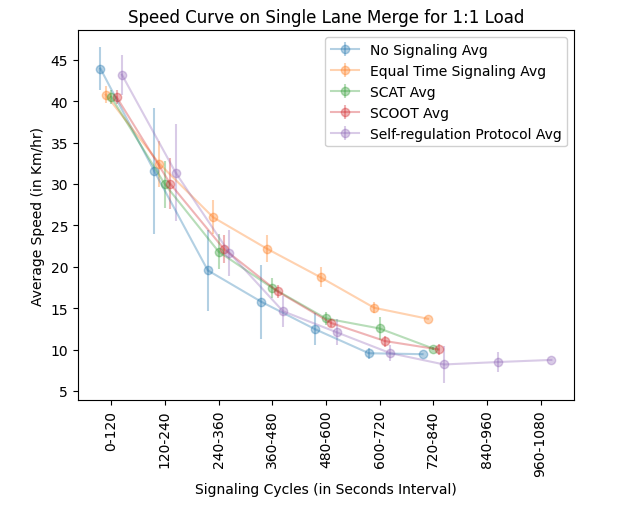}
  \end{subfigure}

  \vskip\baselineskip

  \begin{subfigure}[b]{0.4\linewidth}
    \centering
    \includegraphics[width=\linewidth]{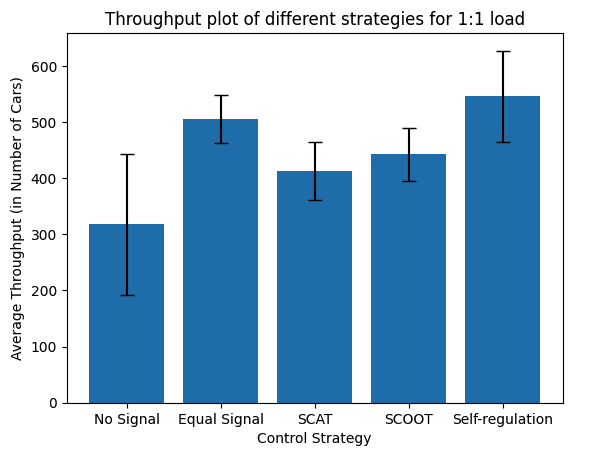}
  \end{subfigure}
  \begin{subfigure}[b]{0.4\linewidth}
    \centering
    \includegraphics[width=\linewidth]{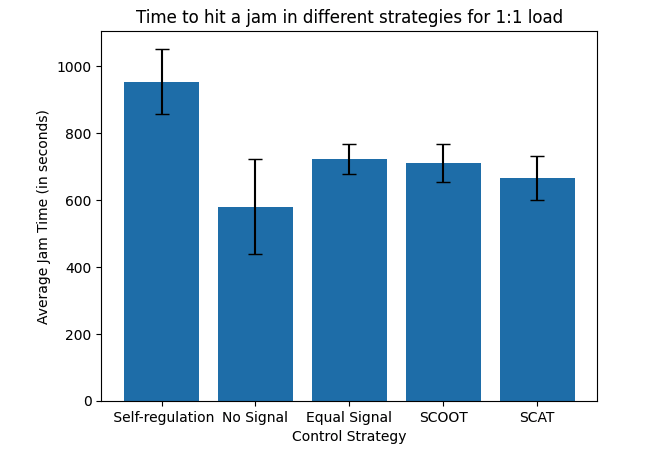}
  \end{subfigure}

  \caption{Flow-Density function plot (top-left) showing the instantaneous flow rate as a function of the queue size. The plot was experimentally derived by simulating a single segment in PTV Vissim. Blue dots are the observed points and the red line is the fitted plot. Performance of different traffic control strategies on single lane merge topology on average speed (top-right), throughput (bottom-left), and time to hit a jam (bottom-right).}
  \label{fig:poc}
\end{figure*}

\paragraph{Physics-Based Proof of Concept}
\label{sec:poc}
Traffic management can be framed as a network performance optimization problem targeting metrics like average travel time, throughput, and fairness, with a key objective of avoiding traffic jams \citep{traffic_curve}. Prior work \citep{dev_engg} has shown that even sub-critical increases in input flow can trigger persistent congestion once critical density is exceeded. Our strategy is thus based on two principles: (1) maintain density below critical thresholds, and (2) maximize flow speed.

To reconcile these, we apply a simplified backpressure-based speed modulation protocol inspired by queue-based routing algorithms from communication networks \citep{tassiulas1990stability, ribeiro2009stochastic, yoo2011experimental, seferoglu2015separation}, and adapted to traffic contexts \citep{ma2020back, chang2020cps, hu2019softpressure}. In a toy 2:1 merge scenario, let $C_\ell(t) = b_\ell(t) \cdot v_\ell(t)$ denote flow on segment $\ell$. The density change on the downstream segment $\ell'$ is given by
\[
    \frac{db_{\ell'}}{dt} = \sum_{\ell_i} C_{\ell_i}(t) - C_{\ell'}(t),
\]
and its remaining capacity before reaching critical density $B^*_{\ell'}$ is $\hat{b}_{\ell'}(t) = B^*_{\ell'} - b_{\ell'}(t)$. If $\delta t$ is the control interval, and $\frac{db_{\ell'}}{dt} \cdot \delta t > \hat{b}_{\ell'}(t)$, input flows are scaled using:
\[
    \alpha = \frac{ \hat{b}_{\ell'}(t)/\delta t + C_{\ell'}(t) }{ \sum_{\ell_i} C_{\ell_i}(t) },
\]
where $\alpha$ applies proportional backpressure via velocity modulation.

Using PTV Vissim, we implemented this protocol and compared it to SCATS, SCOOT\footnote{SCATS and SCOOT were simplified for single intersection and implemented based on original papers.}, equal-time, and no-control baselines on a single-lane merge topology. As shown in Figure~\ref{fig:poc}, backpressure modulation outperformed baselines in throughput and time to jam formation, while intentionally reducing average speed.

However, performance varied with network geometry and driving dynamics. For instance, in single-lane merges, Weidmann 99 and gap acceptance behavior introduced a “gap penalty” at intersections, where vehicles slow down to assess safety before merging. Under high load, this leads to queuing and congestion. We also observed sensitivity to lane configurations, gap timing, and intersection geometry, highlighting the need for more adaptive, scalable strategies.

\section{Methodology}
\label{sec:methodology}
Traffic control in free-flow road networks poses a complex and stochastic optimization challenge. Designing a deterministic speed-modulation policy using first-principles physics is intractable due to nonlinear dynamics, unmodeled driver behavior, and topology-induced variability. Instead, we adopt a reinforcement learning (RL) approach that enables learning from simulation, while embedding traffic-theoretic priors—such as volume-density relationships and gap acceptance theory—into the state and reward design. This hybridization allows for both interpretability and adaptability across varying traffic conditions. Our objective is to learn a control policy that dynamically adjusts segment-wise speed limits to maximize throughput and prevent congestion, without relying on fixed signaling infrastructure.

\paragraph{Environment Setup \& Settings}
To reduce the modeling burden of creating complex traffic networks in PTV Vissim, we developed a semi-automated pipeline that converts real-world road layouts from OpenStreetMap~\citep{openstreetmap} into Vissim-compatible formats. Using Overpass Turbo~\citep{overpass_turbo}, we query and extract a filtered road network based on selected coordinates. The extracted data is then converted to OpenDrive format via \texttt{osm2xodr}~\citep{osm2xodr}, which Vissim can import. While the conversion is not perfect, it significantly reduces the manual effort required to reconstruct large-scale road networks. After generating the network layout, we partition the road into continuous super-segments, each spanning 2–3 kilometers, to serve as control units that capture free-flow traffic characteristics such as volume-density relationships~\citep{nishinari2014traffic,kerner2004three}. OpenStreetMap (OSM) segments are typically much shorter, reflecting changes in road attributes, intersections, or administrative boundaries, and are optimized for geographic accuracy rather than traffic analysis. To construct super-segments, we use sampled vehicle paths from PTV Vissim simulations to identify and group contiguous segments. The RL model operates at the super-segment level, determining speed modulation policies that are fed back to the simulator. Segment-level states within each super-segment are averaged to form a composite state vector, which serves as the model input and enables efficient, interpretable control.

\paragraph{Reinforcement Learning Model Architecture}
We formulate traffic flow control as a Markov Decision Process (MDP), where a centralized agent observes the global traffic state and modulates the maximum allowable velocity on each of $N$ super-segments to optimize throughput and avoid congestion. The MDP is defined by the tuple $\mathcal{M} = ( \mathcal{S}, \mathcal{O}, \mathcal{A}, \mathcal{P}, r, \pi, \gamma )$as follows:

\begin{itemize}[leftmargin=*]
    \item \textbf{State and Observation ($\mathcal{S}, \mathcal{O}$):} At time $t$, each super-segment $i \in \{1, \dots, N\}$ is described by a local state vector $s_i^t \in \mathbb{R}^5$ containing key traffic metrics: vehicle density $\rho_i^t$, average speed $v_i^t$, average gap $g_i^t$, inflow rate $\lambda_i^t$, and outflow rate $\mu_i^t$. These features provide a localized snapshot of congestion, stability, and flow pressure. The global observation is the concatenation $o^t = [s_1^t, \dots, s_N^t] \in \mathbb{R}^{5N}$, forming the input to the agent.
    
    \item \textbf{Action Space ($\mathcal{A}$):} The agent selects an action vector $a^t = [a_1^t, \dots, a_N^t]$, where $a_i^t \in \mathcal{A}_i$ represents the maximum allowable speed (in km/h) for super-segment $i$ during $[t, t+\Delta t]$. We use a fixed, discrete action set $\mathcal{A}_i$ across all segments to promote interpretability, robustness, and compatibility with human-in-the-loop implementations such as in-vehicle advisories.

    \item \textbf{Transition Dynamics ($\mathcal{P}$):} The traffic state evolves stochastically according to $S^{t+1} \sim \mathcal{P}(S^{t+1} \mid S^t, A^t)$, where vehicle interactions are governed by the Weidmann 99 car-following model implemented in the PTV Vissim simulator.

    \item \textbf{Reward Function ($r$):} The reward $r_i^t$ for each super-segment $i$ at time $t$ combines two objectives. First, a penalty is applied if the density exceeds a critical threshold $\rho^* = 0.3$:
    \[
    r^{(1)}_i(t) = \begin{cases}
    -\alpha_d & \text{if } \rho_i^t > \rho^*, \\
    0 & \text{otherwise},
    \end{cases}
    \]
    where $\alpha_d$ is a large penalty constant. Second, the agent is rewarded for maintaining high average speed:
    \[
    r^{(2)}_i(t) = \beta_v \cdot v_i^t,
    \]
    where $\beta_v$ is a positive scaling factor. The total reward for segment $i$ is $r_i^t = r^{(1)}_i(t) + r^{(2)}_i(t)$, and the global reward is $r^t = \sum_{i=1}^N r_i^t$. This design simulates a backpressure-like mechanism, where high-density regions act as implicit pressure signals and speed rewards encourage smooth flow.

    \item \textbf{Objective and Policy ($\pi$, $\gamma$):} The agent aims to learn a policy $\pi: \mathcal{O} \rightarrow \mathcal{A}$ that maximizes the expected cumulative discounted return:
    \[
    \pi^* = \arg\max_\pi \mathbb{E}_\pi \left[ \sum_{t=0}^{T} \gamma^t r^t \right],
    \]
    where $\gamma \in [0,1]$ is the discount factor.
\end{itemize}

To learn the optimal policy, we employ Deep Q-Learning (DQN)~\citep{mnih2015human}. This value-based method is well suited to our centralized setting and discrete action space, offering greater stability and interpretability compared to actor-critic or policy-gradient methods often used in continuous or multi-agent scenarios. The Q-function $Q(o^t, a^t; \theta_n)$ is parameterized by a neural network with weights $\theta_n$, and trained by minimizing the temporal-difference loss:
\[
\mathcal{L}(\theta_n) = \mathbb{E}_{(o^t, a^t, r^t, o^{t'})} \left[ \sum_{i=1}^N \left( r_i^t + \gamma \max_{a'} Q(o^{t'}, a'_i; \theta_{n-1}) - Q(o^t, a_i^t; \theta_n) \right)^2 \right],
\]
where transitions $(o^t, a^t, r^t, o^{t'})$ are sampled from a replay buffer. This formulation supports interpretable learning of control policies that balance throughput and congestion mitigation across dynamic traffic conditions.

\paragraph{Throughput-Optimality under Two-Regime Flow Dynamics}
Let $\lambda_i^t, \mu_i^t$ denote the inflow and outflow in super-segment $i$ at time $t$, and let $f(\rho_i^t)$ be the empirically observed flow-density function. Assume that (i) $f(\rho)$ is unimodal with a unique maximum at critical density $\rho^*$; (ii) The environment evolves as $\rho_i^{t+1} = \rho_i^t + \lambda_i^t - \mu_i^t$; (iii) There is sufficient and persistent inflow: $\lambda_i^t \geq \lambda_{\min} \forall t$; and $\lambda_{\min}$ is large enough that the system would reach $\rho^*$ if outflow were not properly managed. Then, the optimal policy $\pi^*$ that maximizes the expected discounted return, is throughput-optimal in the sense of achieving maximum sustainable flow across the network while avoiding congestion.

To justify this claim, we divide the analysis into two regimes based on the density $\rho_i^t$ in each segment.

\textbf{Free-flow regime ($\rho_i^t < \rho^*$):}  
In this region, the traffic dynamics are dominated by car-following behavior. The vehicle flow is given by
\[
\mu_i^t = \frac{v_i^t}{d_0 + v_i^t T}
\]
where $d_0$ is the standstill distance and $T$ is the headway time. This function strictly increasing in $v_i^t$ for $\rho_i^t < \rho^*$ since
\[
\frac{d\mu_i^t}{dv_i^t} = \frac{d_0}{(d_0+v_i^tT)^2} > 0
\]
Hence, in the free-flow regime flow increases with velocity. Now consider the agent’s reward:
\[
r_i^t = \beta_v \cdot v_i^t \quad \text{(since } \rho_i^t < \rho^* \Rightarrow \text{no penalty)}
\]
So maximizing $r_i^t$ in this region leads to maximizing $\mu_i^t$.

\textbf{Congested regime ($\rho_i^t > \rho^*$):}  
In this region, emergent traffic behaviors dominate (e.g., braking waves), and empirical VDFs (e.g., Figure~\ref{fig:poc}(top-left)) show that:
\[
\frac{d\mu_i^t}{d\rho_i^t} < 0
\]
Additionally, the reward includes a penalty term:
\[
r_i^t = -\alpha_d + \beta_v \cdot v_i^t
\]
The large value of $\alpha_d$ and bounded velocity in the action set ensures that $r_i^t$ is strictly negative. Hence, the agent has incentive to reduce $\rho_i^t$ to exit this regime, but this behavior also maximizes the outflow.

\paragraph{Simulation-Based Training and Transfer Learning}
We train a deep Q-learning model to learn speed modulation policies for self-regulating cars using the PTV Vissim microscopic simulator. The agent controls 12 super-segments, each with three discrete speed options, and is implemented as a two-layer neural network. Every minute, the agent observes a state vector for each super-segment containing density, average speed, gap, inflow, and outflow, and selects speed actions accordingly. Training is performed over simulation episodes of either 1200 or 2500 seconds, using an initial exploration rate of 0.9 that decays to 0.1 (with decay rate 0.02–0.05 per episode), and a discount factor $\gamma = 0.9$ to encourage long-term optimization.

We evaluate two regimes. In the Self-Regulating Cars (SRC) protocol, the model is trained directly on a high-traffic input pattern over 2500 seconds with the action set of (30, 45, or 60 km/h). In the transfer learning variant (SRC-TL), the agent is first trained on a low-traffic input pattern for 1200 seconds to promote exploration across traffic states with the action set of (20, 40, or 60 km/h), then transferred—without fine-tuning—to a different high-traffic pattern for evaluation over 2500 seconds. This setup improves policy robustness by enabling convergence under stable dynamics while ensuring generalization across flow patterns. The use of coarsely discretized actions aids deployability via human-in-the-loop implementations (e.g., mobile advisory systems), and the model is trained solely on instantaneous state and reward signals. This design deliberately avoids temporal input features to ensure transferability across varying demand scenarios, leveraging the consistency of underlying traffic physics such as flow-density and gap acceptance relationships.

\begin{figure}[t]
    \centering
    \begin{subfigure}[b]{0.48\linewidth}
        \centering
        \includegraphics[width=\linewidth]{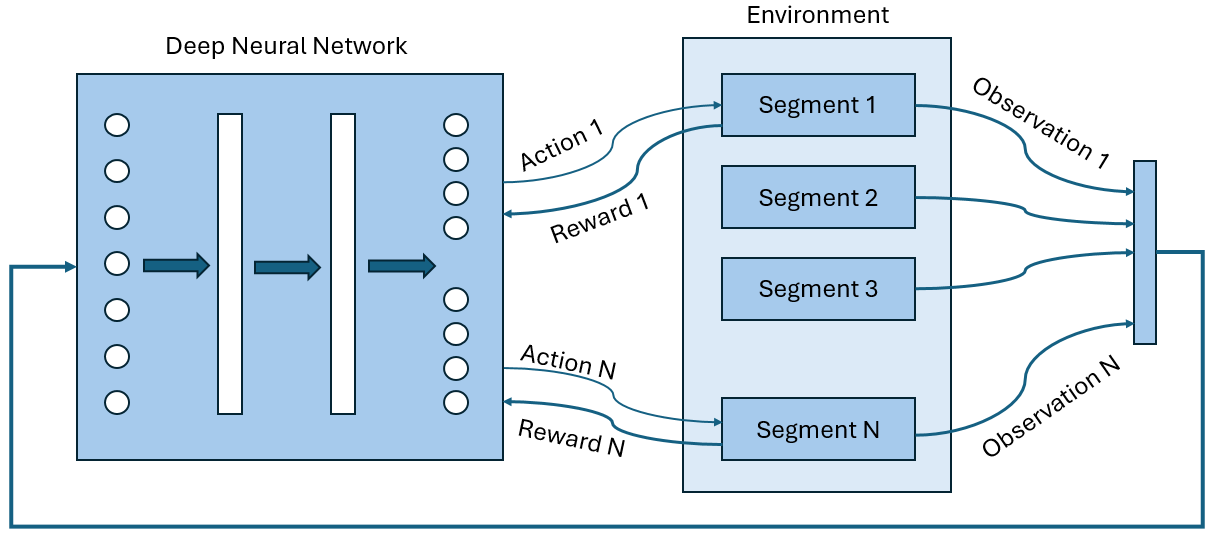}
        \caption{RL-based model to control velocity on different segments of the road network.}
        \label{fig:rl_arch}
    \end{subfigure}
    \hfill
    \begin{subfigure}[b]{0.48\linewidth}
        \centering
        \includegraphics[width=\linewidth]{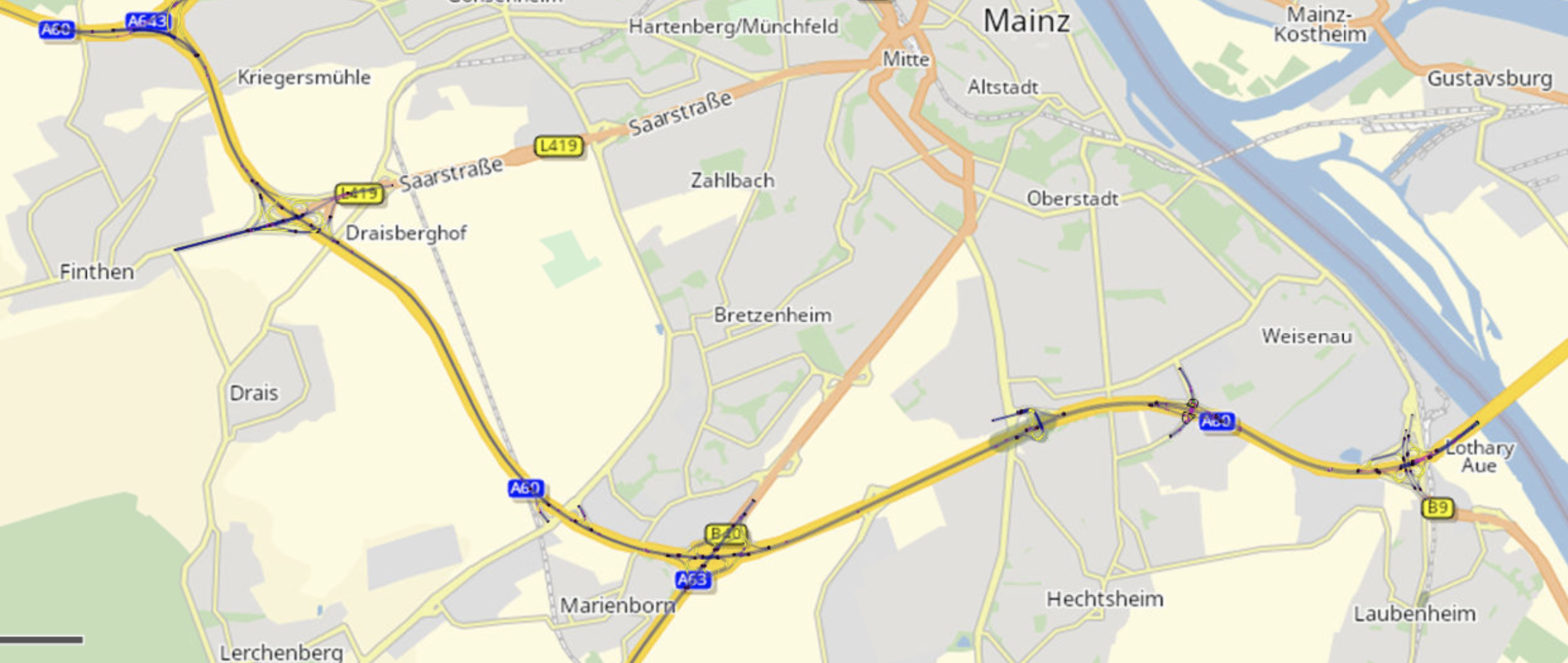}
        \caption{Real-world free-flow road network layout in Mainz, Germany.}
        \label{fig:mainz_layout}
    \end{subfigure}
    \caption{Overview of the reinforcement learning framework and deployment environment.}
    \label{fig:combined_rl_mainz}
\end{figure}

\section{Evaluation}
\label{sec:evaluation}

\paragraph{Evaluation Testbed}
In free-flow road networks, traffic jams rarely occur on uninterrupted straight segments unless capacity changes. While phantom jams \citep{flynn2009traffic} can arise, they are unlikely in our setting where self-driving agents adhere strictly to the Weidmann 99 model. Thus, we focus on applying speed modulation only at network bottlenecks—typically during morning peak hours when suburban traffic converges at city entry points, causing congestion. To study such conditions, we modeled a real-world highway segment in Mainz, Germany (Figure~\ref{fig:mainz_layout}), simulating eastbound rush hour traffic with multiple ramp inflows. The input pattern was designed to reflect realistic cumulative west-to-east flow, where upstream vehicles continue forward and additional traffic merges at successive ramps. The A643 \& A60 junction introduced 3,000–5,000 veh/hr, representing long-distance freeway traffic. L419 (Draisberghof, Saarstraße) added 1,500–3,000 veh/hr from arterial roads. Further east, the A60 \& A63 Marienborn interchange contributed 4,000–6,000 veh/hr from a major highway junction. Finally, the A60 \& Hechtsheim ramp merged 2,500–4,000 veh/hr from urban arterials. No new vehicles entered beyond this point, and all traffic exited via the B9 \& A60 Mainz bridge. This setup allowed Vissim to simulate realistic merging dynamics, congestion buildup, and rising density near the urban core without artificially overloading any single entry point.

\paragraph{Baselines} 
We consider four baselines to compare to our protocol, which are namely, no control, equal split, green wave, and proportional split signaling. By no control, we mean the default behavior of the simulator under the contextual settings. Equal split signaling basically assigns equal amounts of "Green" time to all incoming roads in each cycle. We have taken a cycle time of 120 seconds, with an intergreen time of 5 seconds. In green wave signaling, we add an offset between consecutive signals such that a vehicle going at speed limit would not hit a red light, forming a "green wave". In proportional split signaling, the split of green time is done based on the traffic load in different links. In this case, we allow the intersections with higher net traffic load to have a proportionally higher cycle time. The signaling baselines were implemented by establishing traffic signals at the intersections in the network layout, and they do not manipulate the input traffic in the network, but only modulate how many vehicles can get on the main highway. We want to point out that while the idea of speed modulation has been floated around before, prior works \citep{oh2005dynamic,soriguera2013assessment} have not developed a speed modulation solution for real world road networks that can be evaluated in realistic settings. Hence, we don't have any speed modulation baselines.

\paragraph{Metrics}
We evaluate traffic performance using 14 metrics from PTV Vissim, which we organize into five categories: throughput, flow smoothness, delay, congestion, and emissions.
\begin{itemize}[noitemsep,topsep=0pt,leftmargin=*]
    \item \textbf{Throughput} is captured by \texttt{VEHARR} (vehicles that completed trips), \texttt{DISTTOT} (total distance traveled), \texttt{SPEEDAVG} (average traffic flow speed) and \texttt{DEMANDLATENT} (unfulfilled demand).
    \item \textbf{Flow smoothness} is measured by \texttt{STOPSAVG} and \texttt{STOPSTOT}, representing the average and total number of stops per vehicle, respectively.
    \item \textbf{Delay-related metrics} include \texttt{DELAYAVG}, \texttt{DELAYSTOPAVG}, \texttt{DELAYTOT}, \texttt{DELAYSTOPTOT}, and \texttt{DELAYLATENT}. While we report these for completeness, they are influenced by simulation implementation details and should be interpreted cautiously.
    \item \textbf{Congestion} is quantified via \texttt{VEHACT}, the number of active vehicles remaining in the network, and \texttt{TRAVTMAVG}, which we obtain by normalizing total recorded travel time in vissim (\texttt{TRAVTMTOT}) by total vehicle count (\texttt{VEHARR + VEHACT}). A lower value here reflects less congestion.
    \item \textbf{Emissions} are represented by \texttt{EMISSIONSCO2}, estimated from vehicle speed using the method of \citet{ntziachristos2000speed}. The total calculated emissions are further normalized by multiplying the recorded emissions with (\texttt{VEHARR + VEHACT + DEMANDLATENT}/\texttt{VEHARR + VEHACT}) to account for the latent demand.
\end{itemize}
Taken together, these metrics provide a holistic view of performance of a traffic network.

\renewcommand{\thefootnote}{\fnsymbol{footnote}}
\begin{table*}[t]
    \centering
    \caption{Vehicle network performance measurements for self-regulating cars and baseline traffic management strategies for real-world representative traffic loads. The variance and mean values were calculated over 5 random seeds, with each experiment running for 2500 seconds. Arrow directions show direction of desired outcome.}
    \resizebox{\linewidth}{!}{
    \begin{tabular}{lcccccc}
        \toprule
        \textbf{Metric} & \textbf{No Control} & \textbf{Equal Split} & \textbf{Proportional Split} & \textbf{Green Wave} & \textbf{SRC} & \textbf{SRC (TL)} \\
        \midrule
        \multicolumn{7}{l}{\textbf{Throughput}} \\
        \midrule
        $\uparrow$ VEHARR (\#) & $2227 \pm 19$ & $2002 \pm 22$ & $1873 \pm 10$ & $2045 \pm 17$ & \textbf{2337.8} $\pm$ \textbf{16.42} & 2291.6 $\pm$ 33.78 \\
        \rowcolor{gray!15}
        $\uparrow$ DISTTOT ($10^3$ km) & 12.300 $\pm$ 0.315 & $11.723 \pm 0.086$ & $9.598 \pm 0.083$ & $11.904 \pm 0.050$ & \textbf{13.078} $\pm$ \textbf{0.099} & 12.467 $\pm$ 0.065 \\
        $\uparrow$ SPEEDAVG (km/hr) & 13.16 $\pm$ 0.42 & $11.22 \pm 0.05$ & $10.39 \pm 0.07$ & $10.70 \pm 0.10$ & \textbf{13.23} $\pm$ \textbf{0.16}  & 13.06 $\pm$ 0.06 \\
        \rowcolor{gray!15}
        $\downarrow$ DEMANDLATENT (\#) & $1,722.8 \pm 43.73$ & $1,746.8 \pm 27.53$ & $2,164.8 \pm 32.51$ & 1,570.8 $\pm$ 20.36 & \textbf{1,547.8} $\pm$ \textbf{35.074} & $1,643.8 \pm 57.45$ \\
        \midrule
        \multicolumn{7}{l}{\textbf{Flow Smoothness}} \\
        \midrule
        $\downarrow$ STOPSAVG (\#) & $15.17 \pm 0.34$ & $32.31 \pm 0.50$ & $23.81 \pm 0.52$ & $199.76 \pm 3.19$ & 15.94 $\pm$ 0.96 & \textbf{14.51}$\pm$\textbf{0.32} \\
        \rowcolor{gray!15}
        $\downarrow$ STOPSTOT ($10^3$ \#) & $65.425 \pm 1.711$ & $139.175 \pm 2.059$ & $92.151 \pm 2.345$ & $895.682 \pm 19.05$ & 71.552 $\pm$ 4.165 & \textbf{63.723} $\pm$ \textbf{1.40} \\
        \midrule
        \multicolumn{7}{l}{\textbf{Delay}} \\
        \midrule
        $\downarrow$ DELAYAVG (sec) & $571.42 \pm 6.50$ & $689.11 \pm 3.39$ & $683.65 \pm 9.20$ & $714.00 \pm 6.95$ & 527.20 $\pm$ 9.46 & \textbf{499.36} $\pm$ \textbf{7.96}\footnotemark[1] \\
        \rowcolor{gray!15}
        $\downarrow$ DELAYSTOPAVG (sec) & \textbf{226.26} $\pm$ \textbf{7.31} & $460.62 \pm 3.79$ & $473.28 \pm 9.88$ & $395.73 \pm 5.39$ & 306.05 $\pm$ 11.33 & $251.93 \pm 12.38$\footnotemark[1] \\
        $\downarrow$ DELAYTOT ($10^6$ sec) & $2.464 \pm 0.039$ & $2.968 \pm 0.021$ & $2.646 \pm 0.042$ & $3.201 \pm 0.049$ & 2.366  $\pm$ 0.039 & \textbf{2.193} $\pm$ \textbf{0.030}\footnotemark[1] \\
        \rowcolor{gray!15}
        $\downarrow$ DELAYSTOPTOT ($10^6$ sec) & \textbf{0.976} $\pm$ \textbf{0.034} & $1.984 \pm 0.025$ & $1.832 \pm 0.044$ & $1.774 \pm 0.030$ & 1.373 $\pm$ 0.051 & $1.106 \pm 0.054$\footnotemark[1] \\
        $\downarrow$ DELAYLATENT ($10^6$ sec) & $5.618 \pm 0.107$ & $5.467 \pm 0.076$ & $6.023 \pm 0.088$ & \textbf{5.132} $\pm$ \textbf{0.069} & 5.334 $\pm$ 0.097 & $5.511 \pm 0.117$\footnotemark[1] \\
        \midrule
        \multicolumn{7}{l}{\textbf{Congestion}} \\
        \midrule
        $\downarrow$ VEHACT (\#) & $2085 \pm 38$ & $2305 \pm 32$ & $\textbf{1997} \pm \textbf{11}$ & $ 2438 \pm 55$ & 2150 $\pm$ 27.23 & 2100.2 $\pm$ 14.92 \\
        \rowcolor{gray!15}
        $\downarrow$ TRAVTMAVG ($10^3$ sec/veh) & $\textbf{0.780} \pm \textbf{0.007}$ & $0.873 \pm 0.006$ & $0.859 \pm 0.011$ & $0.894 \pm 0.012$ & $0.793 \pm 0.009$ & $\textbf{0.783} \pm \textbf{0.006}$ \\
        \midrule
        \multicolumn{7}{l}{\textbf{Emissions}} \\
        \midrule
        $\downarrow$ EMISSIONSCO2 (g/m, normalized) & $0.525 \pm 0.013$ & $0.600 \pm 0.001$ & $\textbf{0.513} \pm \textbf{0.005}$ & $3.001 \pm 0.023$ & $0.530 \pm 0.005$ & $0.526 \pm 0.003$ \\
        \bottomrule
    \end{tabular}
    }
    \label{tab:results}
\end{table*}

\footnotetext[1]{Our protocol dynamically adjusts desired speeds, unlike traditional fixed-speed strategies. Since PTV Vissim calculates delay relative to free-flow travel time, which changes with desired speed, a direct numerical comparison may not always be fully representative. To ensure a fair evaluation, we compare delay at equivalent congestion levels and assess other network-wide efficiency metrics.}

\paragraph{Results} Table \ref{tab:results} provides the performance results of our protocol and the baselines on the aforementioned metrics and testbed.  Our self regulating cars protocol (SRC) and transfer learned (SRC(TL)) with a different training input pattern, achieve the best throughput performance across all four metrics (\texttt{VEHARR, DISTTOT, SPEEDAVG, DEMANDLATENT}), aligning with our primary design objective of maximizing network capacity and preventing traffic jams. Our protocol also achieves greatest smoothness in the traffic flow by minimizing the stop frequency (\texttt{STOPSAVG, STOPSTOT}). We also note competitive delay and congestion performance achieved by our protocol with best performance achieved in \texttt{DELAYAVG, DELAYTOT}, and \texttt{TRAVTMAVG}. While the Proportional Split baseline achieves lowest active vehicle numbers (\texttt{VEHACT}), we observed traffic jams and high congestion on entry points to the main highway. Our protocol achieves the best \texttt{VEHACT} with a more uniformly distributed vehicle profile. The \texttt{EMISSIONSC02} numbers are calculated under lots of approximation assumptions, and and the associated variance in calculations is higher than the difference between best emissions and our numbers. Consistent with earlier expectations, our experimental results show that the self-regulating cars protocol performs well even when evaluated on an unseen traffic input pattern. The transferred policy (SRC-TL) continues to optimize key performance metrics, such as throughput and delay, and maintains competitive results across other indicators when compared to both learned and baseline signaling strategies.

\paragraph{Mixed Traffic Ablations}
We conducted ablation experiments on the same network layout, where only a subset of vehicles is controlled by our RL model (25\%, 50\%, and 75\%). All other configuration parameters remained unchanged; however, introducing mixed autonomy led to minor variations in simulation behavior. The results shown in Table \ref{tab:ablation} demonstrate a consistent trend: as the proportion of controlled vehicles increases, network performance improves significantly. Specifically, average speed and throughput increase, while overall delay and congestion-related metrics decrease. These findings reinforce the practical applicability of our method in settings with partial adoption. 

\begin{table}[t]
\centering
\caption{Ablation experiments with mixed traffic inputs. Results show a clear trend of improvement in throughput and flow metrics with increasing fraction of compliant vehicles.}
\vspace{0.15cm}
\resizebox{0.7\linewidth}{!}{
\begin{tabular}{lccc}
\toprule
\textbf{Metric} & \textbf{75\% control} & \textbf{50\% control} & \textbf{25\% control} \\
\midrule
\multicolumn{4}{l}{\textbf{Throughput}} \\
\midrule
$\uparrow$ SPEEDAVG & $13.49 \pm 0.15$ & $13.21 \pm 0.05$ & $12.92 \pm 0.04$ \\
\rowcolor{gray!15}
$\uparrow$ VEHARR & $2270.80 \pm 21.83$ & $2248.67 \pm 56.23$ & $2222.33 \pm 44.61$ \\
$\uparrow$ DISTTOT ($10^3$ km) & $12.61 \pm 0.18$ & $12.29 \pm 0.17$ & $12.04 \pm 0.07$ \\
\rowcolor{gray!15}
$\downarrow$ DEMANDLATENT & $1650.20 \pm 62.86$ & $1720.67 \pm 52.84$ & $1688.00 \pm 54.34$ \\
\midrule
\multicolumn{4}{l}{\textbf{Flow Smoothness}} \\
\midrule
$\downarrow$ STOPSAVG & $14.50 \pm 0.66$ & $14.56 \pm 0.65$ & $14.21 \pm 0.37$ \\
\rowcolor{gray!15}
$\downarrow$ STOPSTOT ($10^3$) & $63.02 \pm 2.53$ & $62.77 \pm 2.41$ & $61.72 \pm 1.19$ \\
\midrule
\multicolumn{4}{l}{\textbf{Delay}} \\
\midrule
$\downarrow$ DELAYAVG & $540.11 \pm 7.26$ & $553.08 \pm 13.83$ & $561.53 \pm 8.43$ \\
\rowcolor{gray!15}
$\downarrow$ DELAYSTOPAVG & $255.82 \pm 10.74$ & $241.66 \pm 10.86$ & $239.65 \pm 10.45$ \\
$\downarrow$ DELAYTOT ($10^6$ s) & $2.35 \pm 0.02$ & $2.38 \pm 0.05$ & $2.44 \pm 0.02$ \\
\rowcolor{gray!15}
$\downarrow$ DELAYSTOPTOT ($10^6$ s) & $1.11 \pm 0.05$ & $1.04 \pm 0.05$ & $1.04 \pm 0.04$ \\
$\downarrow$ DELAYLATENT ($10^6$ s) & $5.53 \pm 0.11$ & $5.64 \pm 0.07$ & $5.65 \pm 0.09$ \\
\midrule
\multicolumn{4}{l}{\textbf{Congestion}} \\
\midrule
$\downarrow$ VEHACT & $2076 \pm 33.31$ & $2063 \pm 41.58$ & $2122 \pm 44.39$ \\
\rowcolor{gray!15}
$\downarrow$ TRAVTMAVG ($10^3$ sec/veh) & $0.775 \pm 0.007$ & $0.777 \pm 0.014$ & $0.771 \pm 0.005$ \\
\bottomrule
\end{tabular}
}
\label{tab:ablation}
\vspace{-0.2in}
\end{table}



\section{Conclusion, Limitations, and Future Work}
\label{sec:conclusion}

We introduced a novel traffic control paradigm based on self-regulating cars—smart vehicles that coordinate to maintain smooth traffic flow in free-flow networks by adjusting speeds according to a physics-informed reinforcement learning policy. Our method achieves comparable or superior performance to existing signaling protocols in high-fidelity real-world simulations, and can be deployed via lightweight integration with in-vehicle navigation systems.

A key limitation of our current setup is the manual specification of road network layouts for simulation, which restricts scalability across different regions. Moreover, the protocol’s effectiveness relies on a sufficient adoption rate of compliant vehicles, and its current design allows non-compliant agents to gain a strategic advantage. In future work, we aim to automate the simulator input pipeline, develop a city scale traffic simulation engine, and extend the control framework to incorporate lane-level decision-making to design mechanisms to mitigate adversarial behavior through incentive-aligned coordination strategies.

\section{Code Availability}
All the relevant code and simulator configuration files can be found on the GitHub repository \url{https://github.com/rohailasim123/Self-Regulating-Cars}.

\section{Acknowledgements}
Ankit Bhardwaj and Lakshminarayanan Subramanian were supported by the NSF Grant (Award Number 2335773)
titled "EAGER: Scalable Climate Modeling using Message-Passing Recurrent Neural Networks". Lakshminarayanan
Subramanian was also funded in part by the NSF Grant (award number OAC-2004572) titled "A Data-informed Framework
for the Representation of Sub-grid Scale Gravity Waves to Improve Climate Prediction".

\bibliographystyle{plainnat}
\bibliography{references}

\begin{thebibliography}{74}
\providecommand{\natexlab}[1]{#1}
\providecommand{\url}[1]{\texttt{#1}}
\expandafter\ifx\csname urlstyle\endcsname\relax
  \providecommand{\doi}[1]{doi: #1}\else
  \providecommand{\doi}{doi: \begingroup \urlstyle{rm}\Url}\fi

\bibitem[Arnold et~al.(1998)]{arnold1998ramp}
ED~Arnold et~al.
\newblock Ramp metering: a review of the literature.
\newblock 1998.

\bibitem[Badger(2013)]{badger2013traffic}
Emily Badger.
\newblock How traffic congestion affects economic growth.
\newblock \emph{Recuperado el}, 18, 2013.

\bibitem[Bhardwaj et~al.(2023)Bhardwaj, Iyer, Ramesh, White, and Subramanian]{dev_engg}
Ankit Bhardwaj, Shiva~R. Iyer, Sriram Ramesh, Jerome White, and Lakshminarayanan Subramanian.
\newblock Understanding sudden traffic jams: From emergence to impact.
\newblock \emph{Development Engineering}, 8:\penalty0 100105, 2023.
\newblock ISSN 2352-7285.
\newblock \doi{https://doi.org/10.1016/j.deveng.2022.100105}.
\newblock URL \url{https://www.sciencedirect.com/science/article/pii/S2352728522000148}.

\bibitem[{California Department of Transportation}(2019)]{pems}
{California Department of Transportation}.
\newblock {PeMS}: {C}alifornia {P}erformance {M}easurement {S}ystem.
\newblock \url{http://pems.dot.ca.gov/}, 2019.
\newblock Accessed: 2024-10-21.

\bibitem[{Cambridge Mobile Telematics}(2025)]{cmt2025}
{Cambridge Mobile Telematics}.
\newblock Cambridge mobile telematics, 2025.
\newblock URL \url{https://www.cmtelematics.com/}.
\newblock Accessed: 2025-05-15.

\bibitem[Center()]{mctrans}
McTrans Center.
\newblock {McT}rans: {T}ransportation software for analysis.
\newblock \url{http://mctrans.ce.ufl.edu/}.
\newblock Accessed: 2024-10-21.

\bibitem[Chang et~al.(2020)Chang, Roy, Zhao, Annaswamy, and Chakraborty]{chang2020cps}
Wanli Chang, Debayan Roy, Shuai Zhao, Anuradha Annaswamy, and Samarjit Chakraborty.
\newblock Cps-oriented modeling and control of traffic signals using adaptive back pressure.
\newblock In \emph{2020 Design, Automation \& Test in Europe Conference \& Exhibition (DATE)}, pages 1686--1691. IEEE, 2020.

\bibitem[Chauhan and Sen(2023)]{chauhan2023reallight}
Sachin Chauhan and Rijurekha Sen.
\newblock Reallight: Drl based intersection control in developing countries without traffic simulators.
\newblock In \emph{NeurIPS 2023 Computational Sustainability: Promises and Pitfalls from Theory to Deployment}, 2023.

\bibitem[Chauhan et~al.(2020)Chauhan, Bansal, and Sen]{chauhan2020ecolight}
Sachin Chauhan, Kashish Bansal, and Rijurekha Sen.
\newblock Ecolight: Intersection control in developing regions under extreme budget and network constraints.
\newblock \emph{Advances in Neural Information Processing Systems}, 33:\penalty0 13027--13037, 2020.

\bibitem[Chauhan and Sen(2024)]{chauhan2024frugallight}
Sachin~Kumar Chauhan and Rijurekha Sen.
\newblock Frugallight: Symmetry-aware cyclic heterogeneous intersection control using deep reinforcement learning with model compression, distillation and domain knowledge.
\newblock \emph{ACM Journal on Computing and Sustainable Societies}, 2\penalty0 (2):\penalty0 1--32, 2024.

\bibitem[Chu et~al.(2019)Chu, Wang, Codec{\`a}, and Li]{chu2019multi}
Tianshu Chu, Jie Wang, Lara Codec{\`a}, and Zhaojian Li.
\newblock Multi-agent deep reinforcement learning for large-scale traffic signal control.
\newblock \emph{IEEE transactions on intelligent transportation systems}, 21\penalty0 (3):\penalty0 1086--1095, 2019.

\bibitem[Croci(2016)]{croci2016urban}
Edoardo Croci.
\newblock Urban road pricing: a comparative study on the experiences of london, stockholm and milan.
\newblock \emph{Transportation Research Procedia}, 14:\penalty0 253--262, 2016.

\bibitem[Downie(2008)]{time_jam}
Andrew Downie.
\newblock The world's worst traffic jams.
\newblock \url{http://content.time.com/time/world/article/0,8599,1733872,00.html}, 2008.

\bibitem[Drake(1967)]{drake1967statistical}
Joseph~S Drake.
\newblock A statistical analysis of speed density hypothesis.
\newblock \emph{HRR}, 154:\penalty0 53--87, 1967.

\bibitem[Flynn et~al.(2009)Flynn, Kasimov, Nave, Rosales, and Seibold]{flynn2009traffic}
Morris~R Flynn, Aslan~R Kasimov, Jean-Christophe Nave, EE~Rosales, and B~Seibold.
\newblock Traffic modeling-phantom traffic jams and traveling jamitons.
\newblock \emph{Traffic}, 8\penalty0 (29):\penalty0 2016, 2009.

\bibitem[for Advanced~Transit and (PATH)(1993)]{path1993}
California~Partners for Advanced~Transit and Highways (PATH).
\newblock A program of research in advanced vehicle control and traffic management for intelligent vehicle-highway systems (ivhs).
\newblock Technical report, Institute of Transportation Studies, University of California, Berkeley, Berkeley, CA, 1993.

\bibitem[Gordon et~al.(1996)Gordon, Reiss, Haenel, Case, French, Mohaddes, Wolcott, et~al.]{gordon1996traffic}
Robert~L Gordon, Robert~A Reiss, Herman Haenel, E~Case, Robert~L French, Abbas Mohaddes, Ronald Wolcott, et~al.
\newblock Traffic control systems handbook.
\newblock Technical report, United States. Federal Highway Administration. Office of Technology Applications, 1996.

\bibitem[Greenshields et~al.(1935)Greenshields, Bibbins, Channing, and Miller]{greenshields1935study}
Bruce~D Greenshields, J~Rowland Bibbins, WS~Channing, and Harvey~H Miller.
\newblock A study of traffic capacity.
\newblock In \emph{Highway research board proceedings}, volume~14, pages 448--477. Washington, DC, 1935.

\bibitem[Hall et~al.(1986)Hall, Allen, and Gunter]{hall1986empirical}
Fred~L Hall, Brian~L Allen, and Margot~A Gunter.
\newblock Empirical analysis of freeway flow-density relationships.
\newblock \emph{Transportation Research Part A: General}, 20\penalty0 (3):\penalty0 197--210, 1986.

\bibitem[Harders(1968)]{harders1968capacity}
J~Harders.
\newblock The capacity of unsignalized urban intersections.
\newblock \emph{Schriftenreihe Strassenbau und Strassenverkehrstechnik}, 76:\penalty0 1968, 1968.

\bibitem[Hu and Smith(2019)]{hu2019softpressure}
Hsu-Chieh Hu and Stephen~F Smith.
\newblock Softpressure: A schedule-driven backpressure algorithm for coping with network congestion.
\newblock \emph{arXiv preprint arXiv:1903.02589}, 2019.

\bibitem[Hunt et~al.(1981)Hunt, Robertson, Bretherton, and Winton]{Hunt1981SCOOTaTR}
P~B Hunt, D.~I. Robertson, R.~D. Bretherton, and R.~I. Winton.
\newblock Scoot-a traffic responsive method of coordinating signals.
\newblock 1981.
\newblock URL \url{https://api.semanticscholar.org/CorpusID:108797003}.

\bibitem[Hunter et~al.(2021)]{hunter2021vissim}
Michael Hunter et~al.
\newblock Vissim simulation guidance.
\newblock Technical report, Georgia. Department of Transportation. Office of Performance-Based~…, 2021.

\bibitem[Hussain et~al.(2016)Hussain, Ghiasi, and Li]{hussain2016freeway}
Omar Hussain, Amir Ghiasi, and Xiaopeng Li.
\newblock Freeway lane management approach in mixed traffic environment with connected autonomous vehicles.
\newblock \emph{arXiv preprint arXiv:1609.02946}, 2016.

\bibitem[Karimi~Shahri et~al.(2019)Karimi~Shahri, Chintamani~Shindgikar, HomChaudhuri, and Ghasemi]{karimi2019optimal}
Pouria Karimi~Shahri, Shubhankar Chintamani~Shindgikar, Baisravan HomChaudhuri, and Amir~H Ghasemi.
\newblock Optimal lane management in heterogeneous traffic network.
\newblock In \emph{Dynamic Systems and Control Conference}, volume 59162, page V003T18A003. American Society of Mechanical Engineers, 2019.

\bibitem[Kerner(2004)]{kerner2004three}
Boris~S Kerner.
\newblock Three-phase traffic theory and highway capacity.
\newblock \emph{Physica A: Statistical Mechanics and its Applications}, 333:\penalty0 379--440, 2004.

\bibitem[Lau(2020)]{google_routing}
Johann Lau.
\newblock Google maps 101: How ai helps predict traffic and determine routes.
\newblock https://blog.google/products/maps/google-maps-101-how-ai-helps-predict-traffic-and-determine-routes/, 2020.

\bibitem[Li et~al.(2021)Li, Yu, Zhang, Dong, and Xu]{li2021network}
Zhenning Li, Hao Yu, Guohui Zhang, Shangjia Dong, and Cheng-Zhong Xu.
\newblock Network-wide traffic signal control optimization using a multi-agent deep reinforcement learning.
\newblock \emph{Transportation Research Part C: Emerging Technologies}, 125:\penalty0 103059, 2021.

\bibitem[Lieu(1999)]{traffic_curve}
Henry Lieu.
\newblock Traffic-flow theory.
\newblock volume~62 of \emph{Public Roads}. Federal Highway Administration, 1999.

\bibitem[Lighthill and Whitham(1955)]{lighthill1955kinematic}
Michael~James Lighthill and Gerald~Beresford Whitham.
\newblock On kinematic waves ii. a theory of traffic flow on long crowded roads.
\newblock \emph{Proceedings of the royal society of london. series a. mathematical and physical sciences}, 229\penalty0 (1178):\penalty0 317--345, 1955.

\bibitem[Lowrie(1982)]{Lowrie1982TheSC}
P.~Lowrie.
\newblock The sydney coordinated adaptive traffic system - principles, methodology, algorithms.
\newblock 1982.
\newblock URL \url{https://api.semanticscholar.org/CorpusID:107741214}.

\bibitem[Ma et~al.(2020)Ma, Xiao, Song, Ma, and Jin]{ma2020back}
Dongfang Ma, Jiawang Xiao, Xiang Song, Xiaolong Ma, and Sheng Jin.
\newblock A back-pressure-based model with fixed phase sequences for traffic signal optimization under oversaturated networks.
\newblock \emph{IEEE Transactions on Intelligent Transportation Systems}, 22\penalty0 (9):\penalty0 5577--5588, 2020.

\bibitem[Mattsson(2008)]{mattsson2008road}
Lars-G{\"o}ran Mattsson.
\newblock Road pricing: Consequences for traffic, congestion and location.
\newblock In \emph{Road pricing, the economy and the environment}, pages 29--48. Springer, 2008.

\bibitem[May et~al.(2005)May, Cayford, Merritt, and Leiman]{bhp}
Adolf~D May, Randall Cayford, Greg Merritt, and Lannon Leiman.
\newblock The {B}erkeley {H}ighway {L}aboratory.
\newblock Path research report, California PATH, University of California, Berkeley, 2005.

\bibitem[May and Milne(2000)]{may2000effects}
Anthony~D May and Dave~S Milne.
\newblock Effects of alternative road pricing systems on network performance.
\newblock \emph{Transportation Research Part A: Policy and Practice}, 34\penalty0 (6):\penalty0 407--436, 2000.

\bibitem[McDonald and Armitage(1978)]{mcdonald1978capacity}
M~McDonald and DJ~Armitage.
\newblock The capacity of roundabouts.
\newblock \emph{Traffic Engineering \& Control}, 19\penalty0 (N10), 1978.

\bibitem[Mnih et~al.(2015)Mnih, Kavukcuoglu, Silver, Rusu, Veness, Bellemare, Graves, Riedmiller, Fidjeland, Ostrovski, et~al.]{mnih2015human}
Volodymyr Mnih, Koray Kavukcuoglu, David Silver, Andrei~A Rusu, Joel Veness, Marc~G Bellemare, Alex Graves, Martin Riedmiller, Andreas~K Fidjeland, Georg Ostrovski, et~al.
\newblock Human-level control through deep reinforcement learning.
\newblock \emph{nature}, 518\penalty0 (7540):\penalty0 529--533, 2015.

\bibitem[Newell(1961)]{newell1961nonlinear}
Gordon~Frank Newell.
\newblock Nonlinear effects in the dynamics of car following.
\newblock \emph{Operations research}, 9\penalty0 (2):\penalty0 209--229, 1961.

\bibitem[Nguyen(2015)]{uber_routing}
Thi Nguyen.
\newblock Eta phone home: How uber engineers an efficient route.
\newblock https://www.uber.com/blog/engineering-routing-engine/, 2015.

\bibitem[Nishinari(2014)]{nishinari2014traffic}
Katsuhiro Nishinari.
\newblock Traffic flow dynamics: Data, models and simulation.
\newblock \emph{Physics Today}, 67\penalty0 (3):\penalty0 54--54, 2014.

\bibitem[Ntziachristos and Samaras(2000)]{ntziachristos2000speed}
Leonidas Ntziachristos and Zissis Samaras.
\newblock Speed-dependent representative emission factors for catalyst passenger cars and influencing parameters.
\newblock \emph{Atmospheric environment}, 34\penalty0 (27):\penalty0 4611--4619, 2000.

\bibitem[OH and Oh(2005)]{oh2005dynamic}
Jun-Seok OH and Cheol Oh.
\newblock Dynamic speed control strategy for freeway traffic congestion management.
\newblock \emph{Journal of the Eastern Asia Society for Transportation Studies}, 6:\penalty0 595--607, 2005.

\bibitem[{OpenStreetMap contributors}()]{openstreetmap}
{OpenStreetMap contributors}.
\newblock Openstreetmap: The free wiki world map.
\newblock \url{https://www.openstreetmap.org}.
\newblock Accessed: 2024-10-21.

\bibitem[Pan et~al.(2021)Pan, Guo, Lam, Zhong, Wang, and He]{pan2021integrated}
Tianlu Pan, Renzhong Guo, William~HK Lam, Renxin Zhong, Weixi Wang, and Biao He.
\newblock Integrated optimal control strategies for freeway traffic mixed with connected automated vehicles: A model-based reinforcement learning approach.
\newblock \emph{Transportation research part C: emerging technologies}, 123:\penalty0 102987, 2021.

\bibitem[Papageorgiou and Kotsialos(2002)]{papageorgiou2002freeway}
Markos Papageorgiou and Apostolos Kotsialos.
\newblock Freeway ramp metering: An overview.
\newblock \emph{IEEE transactions on intelligent transportation systems}, 3\penalty0 (4):\penalty0 271--281, 2002.

\bibitem[Papageorgiou et~al.(2010)Papageorgiou, Papamichail, Messmer, and Wang]{papageorgiou2010traffic}
Markos Papageorgiou, Ioannis Papamichail, Albert Messmer, and Yibing Wang.
\newblock Traffic simulation with metanet.
\newblock \emph{Fundamentals of traffic simulation}, pages 399--430, 2010.

\bibitem[Payne(1973)]{payne1973freeway}
Harold~J Payne.
\newblock Freeway traffic control and surveillance model.
\newblock \emph{Transportation Engineering Journal of ASCE}, 99\penalty0 (4):\penalty0 767--783, 1973.

\bibitem[Peng et~al.(2021)Peng, Keskin, Kulcs{\'a}r, and Wymeersch]{peng2021connected}
Bile Peng, Musa~Furkan Keskin, Bal{\'a}zs Kulcs{\'a}r, and Henk Wymeersch.
\newblock Connected autonomous vehicles for improving mixed traffic efficiency in unsignalized intersections with deep reinforcement learning.
\newblock \emph{Communications in Transportation Research}, 1:\penalty0 100017, 2021.

\bibitem[Prabuchandran et~al.(2014)Prabuchandran, AN, and Bhatnagar]{prabuchandran2014multi}
KJ~Prabuchandran, Hemanth~Kumar AN, and Shalabh Bhatnagar.
\newblock Multi-agent reinforcement learning for traffic signal control.
\newblock In \emph{17th International IEEE Conference on Intelligent Transportation Systems (ITSC)}, pages 2529--2534. IEEE, 2014.

\bibitem[{PTV Group}(2023{\natexlab{a}})]{ptvvissim}
{PTV Group}.
\newblock {PTV Vissim: Microscopic Traffic Flow Simulation Software}.
\newblock \url{https://www.ptvgroup.com}, 2023{\natexlab{a}}.
\newblock Karlsruhe, Germany. Accessed [Insert Date].

\bibitem[{PTV Group}(2023{\natexlab{b}})]{ptvvisum}
{PTV Group}.
\newblock {PTV VISUM: Traffic Planning Software}.
\newblock \url{https://www.ptvgroup.com}, 2023{\natexlab{b}}.
\newblock Karlsruhe, Germany. Accessed [Insert Date].

\bibitem[Raifer()]{overpass_turbo}
Martin Raifer.
\newblock Overpass turbo: A web-based data mining tool for openstreetmap.
\newblock \url{https://overpass-turbo.eu/}.
\newblock Accessed: 2024-10-21.

\bibitem[Ribeiro(2009)]{ribeiro2009stochastic}
Alejandro Ribeiro.
\newblock Stochastic soft backpressure algorithms for routing and scheduling in wireless ad-hoc networks.
\newblock In \emph{2009 3rd IEEE International Workshop on Computational Advances in Multi-Sensor Adaptive Processing (CAMSAP)}, pages 137--140. IEEE, 2009.

\bibitem[Richards(1956)]{richards1956shock}
Paul~I Richards.
\newblock Shock waves on the highway.
\newblock \emph{Operations research}, 4\penalty0 (1):\penalty0 42--51, 1956.

\bibitem[{Samsara Inc.}(2025)]{samsara2025}
{Samsara Inc.}
\newblock Samsara: The connected operations cloud, 2025.
\newblock URL \url{https://www.samsara.com/}.
\newblock Accessed: 2025-05-15.

\bibitem[Seferoglu and Modiano(2015)]{seferoglu2015separation}
Hulya Seferoglu and Eytan Modiano.
\newblock Separation of routing and scheduling in backpressure-based wireless networks.
\newblock \emph{IEEE/ACM Transactions on Networking}, 24\penalty0 (3):\penalty0 1787--1800, 2015.

\bibitem[Shaaban et~al.(2016)Shaaban, Khan, and Hamila]{shaaban2016literature}
Khaled Shaaban, Muhammad~Asif Khan, and Rida Hamila.
\newblock Literature review of advancements in adaptive ramp metering.
\newblock \emph{Procedia Computer Science}, 83:\penalty0 203--211, 2016.

\bibitem[Shi et~al.(2021)Shi, Zhou, Wu, Wang, Lin, and Ran]{shi2021connected}
Haotian Shi, Yang Zhou, Keshu Wu, Xin Wang, Yangxin Lin, and Bin Ran.
\newblock Connected automated vehicle cooperative control with a deep reinforcement learning approach in a mixed traffic environment.
\newblock \emph{Transportation Research Part C: Emerging Technologies}, 133:\penalty0 103421, 2021.

\bibitem[Soriguera et~al.(2013)Soriguera, Torn{\'e}, and Rosas]{soriguera2013assessment}
Francesc Soriguera, Josep~Maria Torn{\'e}, and Dulce Rosas.
\newblock Assessment of dynamic speed limit management on metropolitan freeways.
\newblock \emph{Journal of Intelligent Transportation Systems}, 17\penalty0 (1):\penalty0 78--90, 2013.

\bibitem[Stevanovic et~al.(2009)Stevanovic, Kergaye, and Martin]{stevanovic2009}
Aleksandar Stevanovic, Cameron Kergaye, and Peter Martin.
\newblock Scoot and scats: A closer look into their operations.
\newblock 01 2009.

\bibitem[Tanner(1962)]{tanner1962theoretical}
JC~Tanner.
\newblock A theoretical analysis of delays at an uncontrolled intersection.
\newblock \emph{Biometrika}, 49\penalty0 (1/2):\penalty0 163--170, 1962.

\bibitem[Tassiulas and Ephremides(1990)]{tassiulas1990stability}
Leandros Tassiulas and Anthony Ephremides.
\newblock Stability properties of constrained queueing systems and scheduling policies for maximum throughput in multihop radio networks.
\newblock In \emph{29th IEEE Conference on Decision and Control}, pages 2130--2132. IEEE, 1990.

\bibitem[{Texas A\&M Transportation Institute (TTI)}()]{tti}
{Texas A\&M Transportation Institute (TTI)}.
\newblock {T}exas {T}ransportation {I}nstitute.
\newblock \url{https://tti.tamu.edu/}.
\newblock Accessed: 2024-10-21.

\bibitem[{The osm2xodr contributors}()]{osm2xodr}
{The osm2xodr contributors}.
\newblock osm2xodr: Openstreetmap to opendrive converter.
\newblock \url{https://github.com/JHMeusener/osm2xodr}.
\newblock Accessed: 2024-10-21.

\bibitem[{Transport for New South Wales}(2024)]{scats}
{Transport for New South Wales}.
\newblock \emph{{SCATS} {T}raffic {S}ystem: {S}ydney {C}oordinated {A}daptive {T}raffic {S}ystem}, 2024.
\newblock Accessed: 2024-10-21.

\bibitem[{Transport Research Laboratory (TRL)}(2024)]{scoot}
{Transport Research Laboratory (TRL)}.
\newblock \emph{{SCOOT}: {S}plit {C}ycle {O}ffset {O}ptimisation {T}echnique}, 2024.
\newblock Accessed: 2024-10-21.

\bibitem[Troutbeck and Brilon(1996)]{troutbeck1996unsignalized}
RJ~Troutbeck and W~Brilon.
\newblock Unsignalized intersection theory. in traffic flow theory.
\newblock \emph{Washington, DC: US Federal Highway Administration}, pages 8--1, 1996.

\bibitem[Underwood(1961)]{underwood1961speed}
RT~Underwood.
\newblock Speed, volume, and density relationship. quality and theory of traffic flow, yale bur, 1961.

\bibitem[Vrbani{\'c} et~al.(2021)Vrbani{\'c}, Ivanjko, Mand{\v{z}}uka, and Mileti{\'c}]{vrbanic2021reinforcement}
Filip Vrbani{\'c}, Edouard Ivanjko, Sadko Mand{\v{z}}uka, and Mladen Mileti{\'c}.
\newblock Reinforcement learning based variable speed limit control for mixed traffic flows.
\newblock In \emph{2021 29th Mediterranean Conference on Control and Automation (MED)}, pages 560--565. IEEE, 2021.

\bibitem[Weidmann(1993)]{weidmann1993transporttechnology}
Ulrich Weidmann.
\newblock Transport technology for pedestrians: transport-technical characteristics of pedestrian traffic, literature review.
\newblock \emph{IVT series}, 90, 1993.

\bibitem[Whitham(2011)]{whitham2011linear}
Gerald~Beresford Whitham.
\newblock \emph{Linear and nonlinear waves}.
\newblock John Wiley \& Sons, 2011.

\bibitem[Wiering et~al.(2000)]{wiering2000multi}
Marco~A Wiering et~al.
\newblock Multi-agent reinforcement learning for traffic light control.
\newblock In \emph{Machine Learning: Proceedings of the Seventeenth International Conference (ICML'2000)}, pages 1151--1158, 2000.

\bibitem[Yang et~al.(2020)Yang, Purevjav, and Li]{yang2020marginal}
Jun Yang, Avralt-Od Purevjav, and Shanjun Li.
\newblock The marginal cost of traffic congestion and road pricing: evidence from a natural experiment in beijing.
\newblock \emph{American Economic Journal: Economic Policy}, 12\penalty0 (1):\penalty0 418--453, 2020.

\bibitem[Yoo et~al.(2011)Yoo, Sengul, Merz, and Kim]{yoo2011experimental}
Jae-Yong Yoo, Cigdem Sengul, Ruben Merz, and JongWon Kim.
\newblock Experimental analysis of backpressure scheduling in ieee 802.11 wireless mesh networks.
\newblock In \emph{2011 IEEE International Conference on Communications (ICC)}, pages 1--5. IEEE, 2011.

\end{thebibliography}


\clearpage

\appendix

\section{Tuning PTV Vissim} 
\label{sec:tuning}
For our simulations, we have used PTV Vissim  \citep{ptvvissim}, a state-of-the-art microscopic simulator that uses the Weidmann 99 driving model \citep{hunter2021vissim}. The Weidmann 99 model is a psycho-physical driving behavior model that utilizes discrete thresholds to control driver behavior. Drivers adjust their responses based on factors like gap, speed, or relative speed when these thresholds are reached. This model introduces four driving regimes: In `free-flow regime', drivers independently maintain their desired speed. In `closing-in regime', drivers slow down to match a slower leader's speed and maintain the desired gap. In `following regime', drivers unconsciously mimic the leader's speed and gap. In `emergency braking regime', drivers apply maximum deceleration when the following distance falls below a critical threshold to prevent collisions.

To ensure that the simulated traffic behavior accurately reflects the fundamental diagram of traffic flow, several Weidmann 99 parameters were carefully tuned to reproduce realistic speed-density-flow relationships. The standstill distance (CC0 = 0.5 m) ensures that vehicles are tightly packed in jammed conditions, influencing the jam density of the traffic curve. The gap time distribution (CC1 = 2–3 sec) was set to align with empirical observations of car-following behavior, affecting critical density and maximum throughput. The following distance oscillation (CC2 = 20 m) and distance dependency of oscillation (CC6 = 10 m) help introduce stop-and-go dynamics, ensuring that congestion waves propagate realistically. The threshold for entering following (CC3 = -25 m) defines how aggressively vehicles respond to upstream slowdowns, impacting the onset of congestion. The negative speed difference (CC4 = -0.5) and positive speed difference (CC5 = 0.6) control braking and acceleration tendencies when adapting to varying traffic speeds, influencing capacity drop effects in the fundamental diagram. Lastly, the acceleration parameters (CC7 = 0.5 m/$s^2$, CC8 = 1 m/$s^2$, CC9 = 0.75 m/s² at 80 km/h) were tuned to ensure a smooth transition between free-flow and congested regimes, preventing unrealistic acceleration spikes that could distort the traffic curve. 

As mentioned in \S\ref{sec:traffic_flow_theory}, in free-flow road networks, the traffic stream is modeled with large or even infinite queue sizes, with a relationship between the traffic density and volume of traffic flow given by a non-linear flow-density function which is uni-modal in nature. We have validated this behavior experimentally by simulating a free-flow segment in the PTV Vissim simulator after tuning. The obtained volume-density function plot matches the expected form as shown in Figure \ref{fig:poc} (top-left).



\section{City-scale Signaling Systems}
The simplified versions of SCATS and SCOOT for a single intersection were implemented using the COM interface in PTV VISSIM, following the described methodologies for signal timing allocation at intersections for these systems~\cite{stevanovic2009}.

Both SCATS and SCOOT aim to balance traffic flow at intersections by equalizing saturation levels.  However, they differ in how they define and calculate Degree of Saturation and how they adjust signal timings. SCATS calculates Degree of Saturation (DS) as the ratio of green time used to green time available.  It uses an Incremental Split Selection (ISS) process, adjusting phase splits in small increments each cycle to minimize the DS of the most congested link. These adjustments can be up to ±4\% of the cycle time. SCOOT's Degree of Saturation is calculated as the ratio of traffic demand (Flow Profile) to the product of discharge rate (Saturation Occupancy) and effective green time. As opposed to SCATS which adjusts the signal time in small increments based on the cycle time, SCOOT can adjust the time by absolute intervals of 4 seconds. Based on the DS of each link, SCOOT decides the appropriate time adjustment 5 seconds before each signaling stage transition for the intersection. To prevent rapid fluctuations, SCOOT reverses implemented split changes by 3 seconds before the next cycle starts. For large-scale networks, both protocols take into consideration the current state of multiple intersections within an area that may need to coordinate to improve congestion due to their proximity, and make adjustments accordingly.



\end{document}